\documentclass[10pt, a4paper]{article}

\usepackage[final]{lrec-coling2024} % this is the new style
\usepackage{booktabs}
\usepackage{longtable}
\usepackage{graphicx}
\usepackage{comment}
\usepackage{float}
\usepackage{dblfloatfix}
\usepackage{spverbatim}
\usepackage{xcolor}
\usepackage{caption}  % Include the caption package
\usepackage{arydshln}
\title{A New Massive Multilingual Dataset for \\High-Performance Language Technologies}

\name{Ona de Gibert\textsuperscript{1}, Graeme Nail\textsuperscript{2}, Nikolay Arefyev\textsuperscript{3}, Marta Ba\~{n}\'{o}n\textsuperscript{4}, \\
{\bf Jelmer van der Linde\textsuperscript{2}, Shaoxiong Ji\textsuperscript{1}, Jaume Zaragoza-Bernabeu\textsuperscript{4} } \\
{\bf  Mikko Aulamo\textsuperscript{1}, Gema Ram\'{i}rez-S\'{a}nchez\textsuperscript{4}, Andrey Kutuzov\textsuperscript{3},} \\
{\bf Sampo Pyysalo\textsuperscript{5}, Stephan Oepen\textsuperscript{3} and J\"{o}rg Tiedemann\textsuperscript{1}}}

\address{University of Helsinki, Finland\textsuperscript{1}   University of Edinburgh, UK\textsuperscript{2} University of Oslo, Norway\textsuperscript{3}   \\
Prompsit, Spain\textsuperscript{4} University of Turku, Finland\textsuperscript{5} \\
         \{ona.degibert, shaoxiong.ji, mikko.aulamo, joerg.tiedemann\}@helsinki.fi\textsuperscript{1},\\
         \{graeme.nail, jelmer.vanderlinde\}@ed.ac.uk\textsuperscript{2}, \\
         \{nikolare, andreku, oe\}@ifi.uio.no\textsuperscript{3}\, \\
         \{mbanon, jzaragoza, gramirez\}@prompsit.com\textsuperscript{4},\\
         sampo.pyysalo@utu.fi\textsuperscript{5}}

\abstract{We present the HPLT (High Performance Language Technologies) language resources, a new massive multilingual dataset including both monolingual and bilingual corpora extracted from CommonCrawl and previously unused web crawls from the Internet Archive. We describe our methods for data acquisition, management and processing of large corpora, which rely on open-source software tools and high-performance computing. Our monolingual collection focuses on low- to medium-resourced languages and covers 75 languages and a total of $\approx$ 5.6 trillion word tokens de-duplicated on the document level. Our English-centric parallel corpus is derived from its monolingual counterpart and covers 18 language pairs and more than 96 million aligned sentence pairs with roughly 1.4 billion English tokens. The HPLT language resources are one of the largest open text corpora ever released, providing a great resource for language modeling and machine translation training. We publicly release the corpora, the software, and the tools used in this work.\\
 }

\begin{document}

\maketitleabstract

\section{Introduction}
The development of Large Language Models (LLMs) pre-trained on ever-increasing amounts of text combined with the ongoing advancements in Machine Translation (MT) has made the need for vast amounts of high-quality textual data more pressing than ever. Since the acquisition of large text corpora is a challenge, most works focus on the pre-processing of previously released corpora with new methods, such as more strict textual filters or removal of biased or explicit content. In this work, we present a massive, brand-new dataset for language modeling and MT training based on web crawls produced by the Internet Archive,\footnote{\url{https://archive.org/}} used for the first time at this scale to create multilingual text corpora, and from CommonCrawl.\footnote{\url{https://commoncrawl.org/}}

Under the umbrella of the High Performance Language Technologies (HPLT) project\footnote{\url{https://hplt-project.org/}} \cite{aulamo-etal-2023-hplt}, we obtained access to the web crawls (1.85 PB of data in total at the current stage of the project), downloaded and processed them to create monolingual and parallel corpora with rich metadata: the HPLT language resources. We release the collection under the permissive \texttt{CC0} license\footnote{We do not own any of the text from which these text data has been extracted. We release the data under a specific takedown policy, where any user can ask us to remove their data.} through our project website\footnote{\url{https://hplt-project.org/datasets/}} and OPUS\footnote{\url{https://opus.nlpl.eu/}} \citeplanguageresource{tiedemann2012parallel}. We also publish open-source tools and pipelines used for processing huge web archive data packages so that our real use case can serve as an example for others inside and outside the research community.  Software and tools are released through GitHub.\footnote{\url{https://github.com/hplt-project}}
%The monolingual dataset covers 75 languages and a total of $\approx$ \textcolor{red}{5.6} trillion whitespace-separated word tokens (after de-duplication). Bitexts focus on low- to medium-resourced languages and cover \textcolor{red}{14} language pairs and over \textcolor{red}{36} million aligned documents with roughly \textcolor{red}{6.1} billion tokens altogether.

Our contributions can be summarized as follows:
\begin{itemize}
    \item \textit{monoHPLT}: Monolingual datasets covering 75 languages and over 5.6 trillion tokens.
    \item \textit{biHPLT}: Parallel datasets covering 18 language pairs and over 96 million sentence pairs.
    \item \textit{multiHPLT}: Synthetic datasets obtained by pivoting our parallel datasets through English covering 171 language pairs and 157 million sentence pairs.
    \item \textit{Bitextor} \citep{espla-gomis-etal-2016-bitextors} models: 22 MT models for fast translation and bilingual document alignment covering 9 languages.
    \item \textit{Bicleaner AI} \citep{zaragoza-bernabeu-etal-2022-bicleaner} models: 9 new Bicleaner models for sentence pair scoring.
    \item Scripts and tools for managing, downloading and processing large amounts of web-crawled corpora.
\end{itemize}

The rest of the paper is organized as follows. Section~\ref{sec:related} provides an overview of previous work in constructing corpora for pre-training. Section~\ref{sec:acquisition} describes the acquisition of the presented resources. Section~\ref{sec:datasets} presents in detail the introduced language resources. Finally, Section~\ref{sec:conclusions} concludes our work and discusses future lines of research.

\section{Related Work}
\label{sec:related}

The development of LLMs and highly multilingual MT systems demands large amounts of high-quality data. The scale of training data required by these models makes it effectively impossible to only use curated samples; instead, the common solution to gathering sufficient data is to source it from the Internet. The compilation of text corpora from the Web, both monolingual and bilingual, has been going on for a long time \citep{kilgarriff-et-al-web-as-a-corpus-2003}. While some noteworthy efforts focus on language-specific curated datasets, such as C4 in English \citeplanguageresource{dodge2021documenting} and WuDaoCorpora in Chinese~\citeplanguageresource{WuDaoCorpora}, the current capacity of models in the field has grown, leading to a move towards large multilingual collections.

Regarding monolingual resources, one of the most used sources is CommonCrawl (CC), produced by a non-profit organization that has published a collection of monthly multilingual web snapshots since 2011. Due to its size and noisy nature, there have been multiple efforts at processing CC data to compile cleaned versions: the multilingual OSCAR corpus \citeplanguageresource{suarez2019asynchronous}, as well as the English corpora Pile-CC \citeplanguageresource{gao2020pile}, C4 \citeplanguageresource{dodge2021documenting} and its multilingual counterpart mC4 \citeplanguageresource{xue2021mt5}. 
Other well-known multilingual corpora for language modeling include the recent BigScience ROOTS Corpus \citeplanguageresource{laurenccon2022bigscience}, covering 59 languages from a diverse set of sources, CuturaX~\citeplanguageresource{nguyen2023culturax}, a cleaned multilingual dataset in 167 languages, MADLAD-400~\citeplanguageresource{kudugunta2023madlad}, a large audited dataset in 419 languages, Glot500~\citeplanguageresource{imanigooghari-etal-2023-glot500}, a corpus covering 511 languages, and SERENGETI~\citeplanguageresource{adebara-etal-2023-serengeti}, a dataset in 517 African languages.
\citet{bapna2022building} built a massively multilingual dataset in over 1,500 languages; however, they did not release it publicly. 

% Maybe say something about how important it is to have monolingual resources at document level?

For parallel corpora, the largest publicly available bitext collection is OPUS \citeplanguageresource{tiedemann2012parallel}. The collection includes several large multilingual corpora, such as Paracrawl \citep{banon2020paracrawl}, its current version 9 covers 42 languages and English-centric sentence pairs; CCMATRIX \citeplanguageresource{schwenk2021ccmatrix}, obtained from CC, and the recent NLLB data \citeplanguageresource{costa2022nllb}, which aims at covering as many language pairs as possible.

% Wikimatrix \citeplanguageresource{schwenk2021wikimatrix} with data from Wikipedia,

When dealing with web-crawled corpora, concerns arise regarding the original sources of the data and its level of noisiness. Several works have addressed this issue \citep{kreutzer2022quality,abadji2022towards} and have led researchers to further explore their own datasets and develop new metadata schemes, such as adding genre labels \citep{laippala2022towards,kuzman2023get}, or to include extended annotations such as length, noise and adult content tags \citep{abadji2022towards}. The HPLT language resources also contain additional paragraph-level metadata; see subsection~\ref{subsec:data-mono} for more detail.

\section{From Raw Data to Refined Corpora}
\label{sec:acquisition}

\begin{figure}
    \centering
    \includegraphics[width=\linewidth]{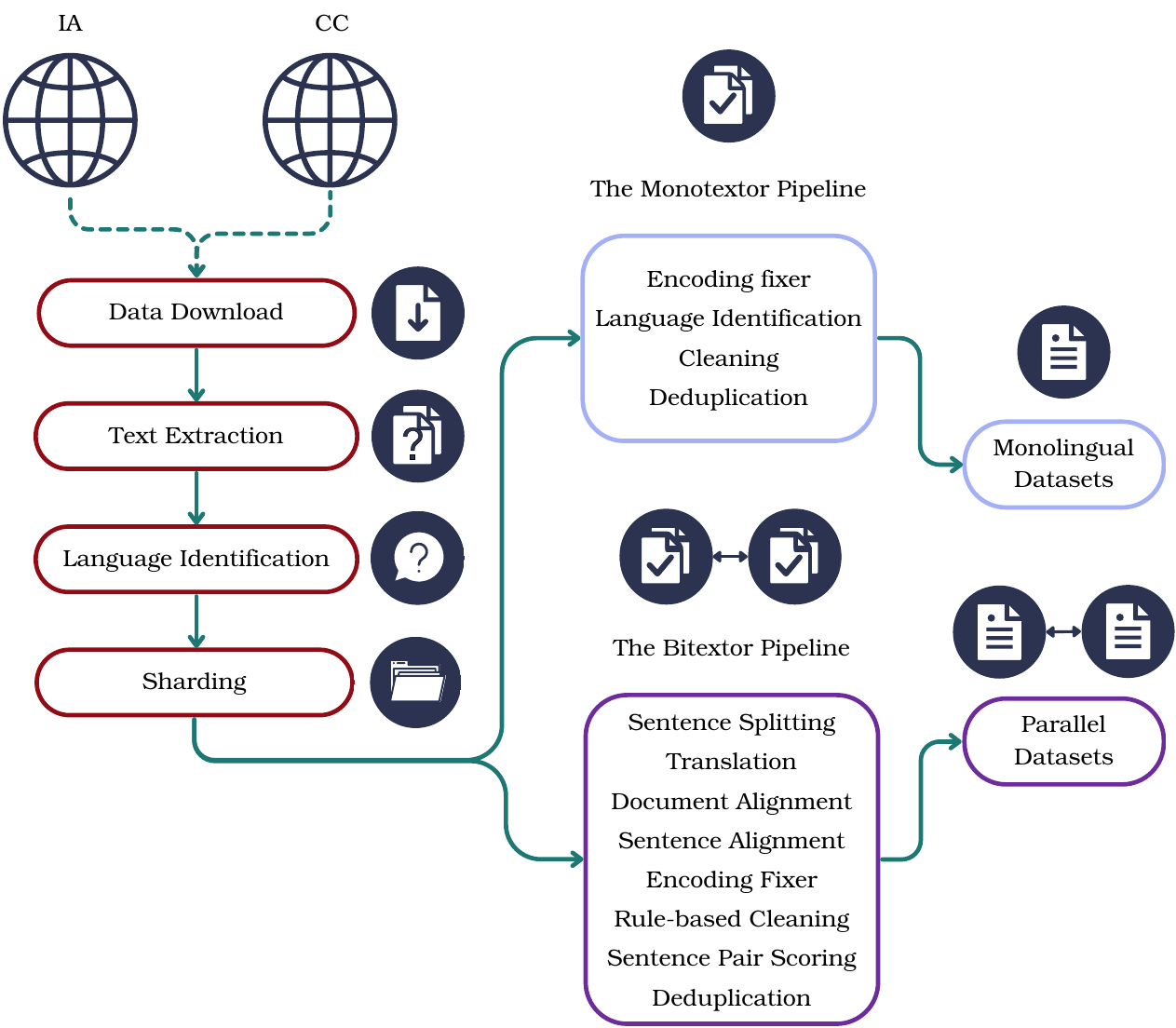}
    \caption{General overview of the HPLT acquisition and processing pipeline.}
    \label{fig:pipeline}
\end{figure}

The management and processing of large datasets both introduce their separate challenges. In this section, we provide a detailed account of the methods, techniques, and considerations employed to collect the raw data and transform it into the corpora presented in Section~\ref{sec:datasets}. A general overview of the pipeline is depicted in Figure~\ref{fig:pipeline}.

\begin{table*}
\centering
\begin{tabular}{lrrrrrl}
\toprule
\textbf{Crawl (collection)}                      & \textbf{CC40} & \textbf{IA WIDE15} & \textbf{IA WIDE16} & \textbf{IA WIDE17} & \textbf{Total} \\
\midrule
\# WARC files                   & 80 000         & 361 431          & 754 143          & 662 381        & 1 857 955   \\
\# files after \texttt{warc2text}        & 384 360        & 1 490 152         & 1 955 584         & 2 403 058 & 6 233 154 \\
Compressed text size, TB        & 8.4           & 19              & 42              & 18    & 87.4           \\
Uncompressed text size, TB     & 18.04         & 38.15           & 130.82          & 43.65        & 230.7    \\
%uncompressed text size for 77 languages, TB     & 18.00         & 38.12           & 103.44          & 43.62            \\
% size reduction after \texttt{warc2text} & 9.88          & 18.84           & 18.29           & 35.61            \\
\# text files                & 127 853        & 495 512          & 977 792          & 798 811      & 2 399 968     \\
%\texttt{warc2text} time, hours               & 23            & 38              & 500             & 48               \\
%\texttt{warc2text} threads              & 245           & 245             & 60              & 245             \\
\bottomrule
\end{tabular}
\caption{Sizes of the raw texts extracted from crawls. `CC' stands for `Common Crawl', `IA' stands for `Internet Archive'.}
\label{tab:wp2_stats_texts}
\end{table*}

\paragraph{Data download} Data acquisition in HPLT relies on two main sources of web crawls: the Internet Archive and Common Crawl. The national High-Performance Computing (HPC) storage resources of Sigma2\footnote{\url{https://www.sigma2.no/data-storage}} and CESNET\footnote{\url{https://www.cesnet.cz/}} were used to download and pre-process web crawls from these two sources. 
The downloading scripts are published in the HPLT git repository.\footnote{\url{https://github.com/hplt-project/ia-download}} These enable parallelized data downloading while automatically verifying and retrying failed downloads after a back-off period. These features are vital for downloading large file collections such as web crawls.

%For the current data release we have downloaded and stored on NIRD two large web crawls from the Internet Archive (IA) named WIDE15 and WIDE17 along with the CC-MAIN-2022-40 (CC40) crawl from Common Crawl. These crawls occupy a total of 1082 TB on NIRD. The download speed varied between 9 and 33 TB/day. On CESNET, we downloaded a third crawl from Internet Archive,  WIDE16, which accounts for 768 TB. The download speed was 17.5TB/day. The web crawls are available in the WARC (Web Archive) format\footnote{\url{https://www.iso.org/standard/68004.html}}.

For the current data release, we have downloaded three large web crawls from the Internet Archive (IA) named WIDE15, WIDE16 and WIDE17, along with the CC-MAIN-2022-40 (CC40) crawl from Common Crawl. These crawls occupy a total of 1850 TB and are stored in WARC (Web Archive) format\footnote{\url{https://www.iso.org/standard/68004.html}}. More data will be made available in the future releases.

\paragraph{Text Extraction} WARC files contain many types of data besides written text: images, sound, video, etc. In order to extract raw texts and conduct preliminary language identification, the downloaded crawls were processed by the \texttt{warc2text} tool from the Bitextor pipeline.\footnote{\url{https://github.com/bitextor/warc2text}} \texttt{warc2text} finds documents containing text in some natural language and does fast preliminary filtering of undesirable documents based on their URL or HTML tags. More thorough filtering happens at the next stages. From the remaining documents, it extracts raw unformatted text and performs initial, document-level language detection. Running whitespace is normalized, and paragraph-like segments, as defined by HTML block elements (\texttt{<p>}, \texttt{<ul>}, \texttt{<ol>}, etc.) are encoded as newlines in this raw text. The output of \texttt{warc2text} consists of compressed base64-encoded raw texts along with the URLs of the original web pages these texts originate from. This data is grouped into directories by language, which is detected using the CLD2 language classifier\footnote{\url{https://github.com/CLD2Owners/cld2}}. Table~\ref{tab:wp2_stats_texts} presents summary statistics for the crawls, showing that out of the four sources, WIDE16 produces by far the largest amount of text. In this step, we obtained 87.4 TB of compressed or 230.7 TB of uncompressed text in total.

After text extraction, we selected 77 languages with the highest amount of obtained raw text for this data release. We plan to add more languages in the following releases. 
% Figures \ref{fig:text_bytes_props1} and \ref{fig:text_bytes_props2} show the volumes of texts extracted by \texttt{warc2text} for each language.
The volume of uncompressed text obtained differs significantly across languages, from 2.2 GB for text classified by CLD2 as Esperanto to 77.5 TB for English, while the number of documents has a minimum of 314K for Pashto and a maximum of 12.8B for English. 
% Figures~\ref{fig:text_bytes_props1} and \ref{fig:text_bytes_props2} show the distribution of extracted text per crawl source for each language.
For most languages, the majority of texts come from the largest crawl, WIDE16; however, for Chinese, the main source is WIDE17, while Esperanto, Basque, and Nepali primarily come from CC40, despite it containing only a fifth of the text that WIDE16 does. Thus, a combination of different crawls, including small ones, seems to be beneficial for good coverage of different languages. The source crawl distribution per language can be consulted in Appendix~A.

\begin{comment}
    
\begin{figure}[ht!]
\centering
\includegraphics[width=\linewidth]{images/langstats/text_bytes_prop1.pdf}
\caption{Proportions of text volume in bytes coming from each crawl, part 1.}
\label{fig:text_bytes_props1}
\end{figure}

\begin{figure}[ht!]
\centering
\includegraphics[width=\linewidth]{images/langstats/text_bytes_prop2.pdf}
\caption{Proportions of text volume in bytes coming from each crawl, part 2.}
\label{fig:text_bytes_props2}
\end{figure}
\end{comment}

\paragraph{Language Identification}

The preliminary per-document language identification employs the CLD2 language identifier described above as the fastest solution. It is conducted as a part of \texttt{warc2text} processing. However, at a later stage of data processing (see below), we use FastSpell\footnote{\url{https://github.com/mbanon/fastspell}} \cite{banon2024fast} for more accurate language identification at the level of paragraph-like segments.

\paragraph{Sharding} To better deal with the amount of data to be processed, we organise the raw text records into 256 shards. The Bitextor pipeline identifies parallel text within a single shard, \footnote{\url{https://github.com/bitextor/bitextor/blob/master/docs/CONFIG.md\#preprocessing-and-sharding}} and therefore, records are placed into shards by their domain name, excluding the top-level domain, to increase the likelihood of matches. Since the distribution of data is not uniform across shards, we batch the data into equally sized chunks for each shard to help balance the computational requirements. For monolingual text extraction, the division among shards is ignored.

%Records are grouped in the same batches based on a hash derived from the domain name of the URL of the record, modulo 256, resulting in 256 shards, each containing one or many batches. This grouping of records by domain is done to help with the identification of parallel texts. The assumption is that pages that are translations of each other are likely to be using the same domain name. To account for translations being hosted on their own top-level domain, this part of the domain name is ignored when generating the hash. The top-level domain of the URL is determined using the Public Suffix List\footnote{\url{https://publicsuffix.org}}. F

After these steps are complete, the monolingual and bilingual text processing pipelines go on separately, as described below. The following steps have been performed on the EuroHPC cluster LUMI\footnote{\url{https://www.lumi-supercomputer.eu/}}.

\subsection{Monolingual Text Processing}

After the sharding step, we process the monolingual extracted text with cleaning tools in order to perform fixes at character level and to enrich the corpora with additional metadata that can be used to produce filtered versions for different applications. 

\paragraph{The Monotextor Pipeline}  To be able to process 124TB of compressed text and scale across many compute nodes in an HPC cluster, a new pipeline based on Slurm scripts using the existing Monotextor tool was developed.\footnote{\url{https://github.com/hplt-project/monotextor-slurm}} The pipeline performs the following steps:
\begin{enumerate}
    \item \textbf{TSV Formatting}: for each shard, a tab-separated file is created where each line contains a document URL, a text paragraph, and a collection name. The file is split into batches of equal amounts of uncompressed text to balance subsequent processing jobs. For the following steps, each batch file is processed in parallel across the number of lines with GNU Parallel \citep{tange_2023_7558957}.
    \item \textbf{Monofixer}\footnote{\url{https://github.com/bitextor/bifixer}}:  every line, containing a paragraph-like segment of text, is processed by the character and encoding fixer, including fixing mojibake (encoding errors), unescaping HTML entities and removing HTML tags.
    \item \textbf{FastSpell}: We perform language identification at paragraph-level in two steps. First, each paragraph receives a language tag given by fastText. Then, we refine the language identification using Hunspell dictionaries for improved precision. We check the paragraph for spelling errors with Hunspell dictionaries based on a list of similar languages\footnote{\url{https://github.com/mbanon/fastspell/blob/main/src/fastspell/config/similar.yaml}} to the one identified by fastText. The language whose dictionary produces the least spelling errors is the final prediction.
    \item \textbf{Monocleaner}\footnote{\url{https://github.com/bitextor/monocleaner}}: each paragraph is also assigned a fluency score, computed with a 7-gram modified Knesser-Ney character language model. Each language model (one per language) is trained on samples of about 200,000 sentences mostly coming from the monolingual part of OPUS corpora. % \citeplanguageresource{tiedemann2012parallel}.
    Only corpora coming from non-web-crawled data and languages that have not been automatically identified are chosen. Data from Wikipedia dumps are used for languages that do not have enough data in OPUS. This fluency score can be used to estimate the `quality' of paragraphs in the document, allowing to filter out noise that may be detrimental for training language models.
    \item \textbf{JSON formatting}: Finally, each batch tab-separated file is converted to JSON-lines (JSONL) format.
\end{enumerate}

\paragraph{Language Mappings} In addition to the processing described above, some modifications to how languages are stored after WARC text extraction have been made:
\begin{itemize}
    \item CLD2 uses old Hebrew `\texttt{iw}' language code, so it has been renamed to use the official `\texttt{he}'.\footnote{\url{https://www.loc.gov/standards/iso639-2/php/langcodes_name.php?iso_639_1=he}}
    \item Norwegian Bokmål is identified as `\texttt{no}' by CLD2, so it has been changed to `\texttt{nb}' to avoid possible confusion, as `\texttt{no}' may also refer to all the Norwegian variants, not only Bokmål. % alos its a pain when you write "no" in yaml without quotes ~ Jelmer :D
    \item For consistency with the rest of the languages, where we are not separating by writing script, traditional and simplified Chinese (`\texttt{zh-Hant}' and `\texttt{zh-Hans}') have been merged into `\texttt{zh}'.
    \item For the monolingual collection, Serbo-Croatian languages (Bosnian `\texttt{bs}', Croatian `\texttt{hr}' and Serbian `\texttt{sr}') have been merged under the `\texttt{hbs}' code. Because of their mutual intelligibility, these languages are often mixed up with each other during language identification.
\end{itemize}

\noindent
This process leaves \textbf{75 languages} for the monolingual data release.

\paragraph{De-duplication}
De-duplication of training corpora is of utmost importance, especially in the case of web-crawled text collections, which can often contain multiple copies of the same text appearing on different web pages.
At the same time, overly aggressive de-duplication can lead to biased corpora, which are not representative of frequency patterns in the corresponding languages anymore. The datasets we release aim to allow end users to decide whether they would like to apply any additional pre-processing. 

For these reasons, we limited ourselves to removing near-duplicates \textbf{on the document level}, using a variation of the MinHash algorithm \citep{broder2000identifying}. This removed approximately 70-80\% of the original data for high-resource languages and 40\% for low-resource languages. In total, after de-duplication, the monolingual dataset was reduced to nearly a third of the size of the original (from 21 TB to 7.5 TB), but at the same time much more balanced and suitable for training language models.
Note that we also release the data \textit{before de-duplication} to preserve the possibility to reproduce or refine our de-duplication pipeline.

The final statistics and data format of the \mbox{monoHPLT} corpora
are described in subsection~\ref{subsec:data-mono} below.

\subsection{Bitext Extraction}

The sharded monolingual data is the input to the bitext extraction pipeline used to create parallel corpora. We rely on previous experience from the ParaCrawl\footnote{\url{https://www.paracrawl.eu/}} and MaCoCu projects\footnote{\url{https://macocu.eu/}} and adjust tools and procedures from the Bitextor pipeline\footnote{\url{https://github.com/paracrawl/cirrus-scripts/tree/lumi}} according to the needs and languages in the HPLT setup. 

In this release, we focus on English-centric data as we expect the largest potential outcome of parallel data from the alignment to English. Furthermore, the Bitextor pipeline relies on automatic document translation in one of the steps and the performance of translations into English is more reliable than translations into other languages, especially for lesser-resourced languages. The initial release covers 18 language pairs with a strong focus on lesser-resourced languages and a few non-European languages to increase the diversity of parallel data available for MT development.

\paragraph{The Bitextor Pipeline} The bitext extraction pipeline is based on Bitextor.\footnote{\url{https://github.com/bitextor/bitextor}} 
We extended scripts developed for ParaCrawl \cite{banon2020paracrawl} for scheduling and workflow automation to meet our needs. We adjusted and further developed the pipeline for the needs of HPLT on the LUMI supercomputer. Mining bilingual sentence pairs using this pipeline and English as one of the languages consists of the following processing steps for each language pair:

\begin{enumerate}
    \item \textbf{Sentence Splitting}: splits the documents into sentences using a language-specific sentence splitter. When there is no language-specific sentence splitter, default to English.
    \item \textbf{Translation}: translate the non-English sentences into English for document alignment. For these steps, we needed to develop fast MT models described below.  MarianNMT~\citep{junczys2018marian} was used for automatic translation and adapted to work with the AMD GPUs available on LUMI.\footnote{\url{https://github.com/hplt-project/lumi-marian}}
    \item \textbf{Document Matching}: match English and translated documents using TF/IDF. Code was improved\footnote{\url{https://github.com/hplt-project/document-aligner}} to work with less memory to better handle larger batch sizes. 
    \item \textbf{Sentence Alignment}: match English and translated sentences in the matched documents using Bleualign\footnote{\url{https://github.com/bitextor/bleualign-cpp}} \citep{sennrich2010mt}, which relies on the English translated sentence and the original English sentence.
    \item \textbf{Bifixer} \citep{prompsit:2020:EAMT}: fix encoding and orthographic issues, similar to Monofixer for monolingual text data.
    \item \textbf{Bicleaner-hardrules} \citep{prompsit:2020:EAMT}: remove noisy sentence pairs for obvious noise based on rules, poor language identified using FastSpell and vulgar language based on specific language modelling. 
    \item \textbf{Bicleaner AI} \citep{zaragoza-bernabeu-etal-2022-bicleaner}: score sentence pairs to indicate whether they are mutual translations (with a value near to 1) or not (with a value near to 0). We keep sentences whose Bicleaner score is above 0.5.
    \item \textbf{Reduce}: collect and concatenate all data across shards and collections. In our case, this is a combination of all the data extracted from CC-40, WIDE15, WIDE16 and WIDE17.
    \item \textbf{De-duplication and TMX Formatting}: the final step generates a TMX file. In this step, the sentence pairs are de-duplicated, ignoring differences in punctuation. The source URLs are retained so that a single sentence pair can have multiple URLs, identifying all the documents that it occurred in.
\end{enumerate}

\subsubsection{Bitextor Models}
Document matching in the Bitextor pipeline requires the translation of one language into the other in order to use efficient monolingual matching strategies to find parallel documents in the vast space of extracted texts. This requires efficient translation models to make it computationally feasible to process data at the scale involved in this work. Bitextor already supports a number of languages from prior work, but their coverage is limited. OPUS-MT\footnote{\url{https://github.com/Helsinki-NLP/OPUS-MT}} provides additional resources in terms of pre-trained models that can be employed directly for translation or for distillation, as detailed below.

For this data release, we trained new efficient student MT models to enable the extraction of additional language pairs. We adopted larger transformer-based MT systems as teacher models and distilled knowledge from the teacher to train student models and improve efficiency via sequence-level knowledge distillation \citep{kim2016}. This technique allows the student model to learn from the teacher model to create a model of comparable quality but improved throughput thanks to its smaller size. We trained two different-size student models, \texttt{base} and \texttt{tiny}, to cover different quality-throughput requirements. We trained models for languages including \texttt{ar}, \texttt{ca}, \texttt{eu}, \texttt{gl}, \texttt{hi}, \texttt{jp}, \texttt{sw}, \texttt{vi}, and \texttt{zh} (in both simplified and traditional scripts, as well as a joint model). We release the student models under our GitHub repository\footnote{\url{https://github.com/hplt-project/bitextor-mt-models}} .

\begin{figure*}[t!]
    \centering
    \includegraphics[width=0.7\textwidth]{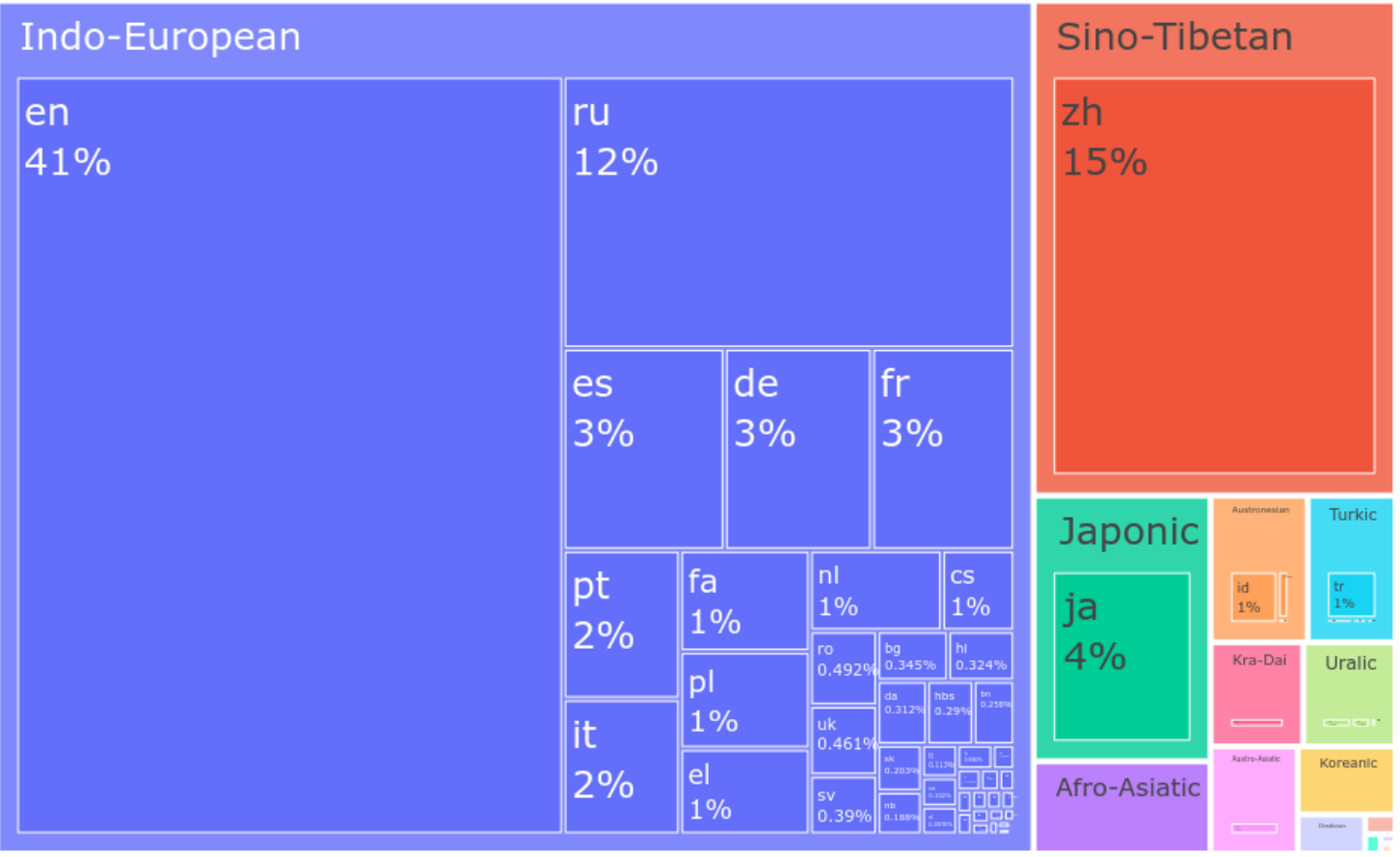}
    \caption{Size distribution for the monolingual corpora, organized by language family and language. The volume of texts ranges from 1.0 GB for text classified by CLD2 as Esperanto to 20.3 TB for English, accounting for 41\% of the whole collection.}
    \label{fig:mono-lang-treemap}
\end{figure*}

\subsubsection{Bicleaner AI Models}

For bilingual data filtering, we use Bicleaner AI, which aims to detect noisy sentence pairs in a parallel corpus. It indicates the likelihood of a pair of sentences being mutual translations (with a value near to 1) or not (with a value near to 0). Sentence pairs considered very noisy are scored 0.

Although there are already Bicleaner models available, we trained new Bicleaner models for the language pairs that we include in this release: \texttt{ar}, \texttt{eu}, \texttt{gl}, \texttt{he}, \texttt{hi}, \texttt{jp}, \texttt{sw}, \texttt{vi}, and \texttt{zh}.
We have increased the total amount of language pairs available from 36 to 45,\footnote{\url{https://huggingface.co/models?other=bicleaner-ai}}
also including many changes and improvements to the tool since version 1.0.1 made for ParaCrawl\footnote{\url{https://github.com/bitextor/bicleaner-ai/blob/v2.3.2/CHANGELOG.md}}. We make all of the newly introduced models available for download.\footnote{\url{https://github.com/bitextor/bicleaner-ai/\#download-a-model}}

\begin{table*}[t!]
\small
\centering
\setlength{\tabcolsep}{2pt}
\begin{tabular}{lrrrrrr}
\toprule
           & \multicolumn{2}{c}{\textbf{Raw}}                                                                   & \multicolumn{2}{c}{\textbf{Filtered}} & \multicolumn{2}{c}{\textbf{De-duplicated}}                                                                                                 \\
           \textbf{Language Pair} &  \multicolumn{1}{r}{\textbf{\# Segments}} & \multicolumn{1}{r}{\textbf{\# Tokens}} & \multicolumn{1}{r}{\textbf{\# Segments}} & \multicolumn{1}{r}{\textbf{\# Tokens}} & \multicolumn{1}{r}{\textbf{\# Segments}} & \multicolumn{1}{r}{\textbf{\# Tokens}} \\
\midrule
Norwegian (nn)  &  28 701 601  &  496 496 331  &  649 435  &  6 308 500  &  132 538  &  2 082 878\\
Bosnian* (bs)  &  26 998 901  &  521 626 621  &  1 426 670  &  12 439 348  &  240 012  &  2 705 525\\
Basque (eu)  &  20 830 243  &  400 262 771  &  3 087 453  &  31 739 210  &  610 687  &  9 964 617\\
Maltese (mt)  &  135 103 434  &  2 820 798 439  &  9 170 421  &  133 140 189  &  854 820  &  18 819 145\\
Gaelic (ga)  &  101 001 090  &  2 013 971 167  &  15 644 170  &  144 323 574  &  994 746  &  16 327 484\\
\hdashline
Galician (gl)  &  56 101 411  &  1 015 559 754  &  5 789 361  &  49 604 655  &  1 063 103  &  13 904 758\\
Macedonian (mk)  &  91 293 129  &  1 868 196 128  &  20 474 476  &  221 370 998  &  1 139 051  &  18 562 461\\
Albanian (sq)  &  253 098 546  &  5 819 014 143  &  16 729 596  &  144 732 656  &  1 655 958  &  25 831 054\\
Swahili (sw)  &  247 557 313  &  5 746 490 123  &  24 448 577  &  209 062 077  &  1 710 205  &  20 039 612\\
Icelandic (is)  &  170 419 019  &  3 266 074 902  &  28 149 571  &  262 486 823  &  2 148 854  &  29 493 241\\
Serbian* (sr)  &  754 277 462  &  14 249 438 714  &  60 482 286  &  586 909 655  &  4 643 025  &  67 063 293\\
Chinese (zh)  &  530 119 983  &  9 162 123 041  &  47 852 076  &  510 404 638  &  5 306 570  &  83 811 653\\
Estonian (et)  &  865 431 226  &  15 476 948 993  &  72 976 009  &  752 767 471  &  6 089 791  &  95 943 562\\
Catalan (ca)  &  402 492 626  &  8 034 120 323  &  88 434 510  &  882 436 335  &  8 905 889  &  141 859 163\\
Croatian* (hr)  &  895 785 142  &  16 565 285 999  &  128 145 132  &  1 165 895 906  &  9 310 275  &  138 360 666\\
Hindi (hi)  &  1 043 856 525  &  19 246 270 565  &  117 341 153  &  996 036 740  &  12 043 069  &  165 139 713\\
Arabic (ar)  &  1 545 148 805  &  33 199 212 426  &  277 864 501  &  2 307 727 128  &  14 645 128  &  239 377 462\\
Finnish (fi)  &  3 826 974 191  &  65 312 092 463  &  495 310 671  &  4 186 819 006  &  25 176 462  &  338 063 309\\
\midrule
Total  &  10 995 190 647  &  205 213 982 903  &  1 413 976 068  &  12 604 204 909  &  96 670 183  &  1 427 349 596\\
\bottomrule
\end{tabular}
\caption{Statistics on the extracted bitexts without filtering (Raw), after cleaning (Filtered) and after de-duplication (De-duplicated) ordered by available clean de-duplicated segments. All statistics are measured from the English side of each language pair. The symbol * indicates that a joint Bicleaner AI model has been used for processing those languages. The dashed line marks the boundary of 1 million clean and de-duplicated segments, which is often used as a threshold to distinguish low-resource and higher-resource languages.}
\label{tab:extracted_bitexts}
\end{table*}

\section{The HPLT Language Resources}
We next present the HPLT language resources, a new massive multilingual dataset covering 75 monolingual and 18 bitext corpora. We release all our collections under the permissive \texttt{CC0} license through our project website and OPUS.

\label{sec:datasets}
\subsection{Monolingual Datasets}
\label{subsec:data-mono}
Our monolingual collection covers 75 languages. We include high-resource languages such as English (en), Chinese (zh), Russian (ru) and Japanese (jp), as well as low-resource ones such as Esperanto (eo), Pashto (ps), Tatar (tt) and Welsh (cy). The full statistics for all languages are presented in Appendix~B.

%: Esperanto (eo), Pashto (ps), Tatar (tt), Welsh (cy), Kyrgyz (ky), Somali (so), Irish (ga), Norwegian (nn), Basque (eu), Swahili (sw), Maltese (mt), Gujarati (gu), Uzbek (uz), Punjabi (pa), Galician (gl), Kannada (kn), Icelandic (is), Tagalog (tl), Sinhalese (si), Macedonian (mk), Mongolian (mn), Marathi (mr), Afrikaans (af), Kazakh (kk), Armenian (hy), Nepali (ne), Telugu (te), Urdu (ur), Belarusian (be), Malayalam (ml), Burmese (my), Latin (la), Georgian (ka), Azerbaijani (az), Albanian (sq), Latvian (lv), Estonian (et), Slovenian (sl), Catalan (ca), Lithuanian (lt), Tamil (ta), Norwegian Bokmål (nb), Slovak (sk), Malay (ms), Bengali (bn), Finnish (fi), Serbo-Croatian (hbs), Hebrew (he), Danish (da), Hindi (hi), Bulgarian (bg), Swedish (sv), Hungarian (hu), Ukrainian (uk), Romanian (ro), Korean (ko), Czech (cs), Vietnamese (vi), Thai (th), Indonesian (id), Turkish (tr), Dutch (nl), Arabic (ar), Greek (el), Polish (pl), Persian (fa), Italian (it), Portuguese (pt), French (fr), German (de), Spanish (es), Japanese (ja), Russian (ru), Chinese (zh), English (en). 
In total, after de-duplication, we release a collection of 5.25 billion documents (approximately corresponding to web pages), totaling 50.1 TB of uncompressed texts and approximately 5.6 trillion whitespace-separated word tokens.  Figure~\ref{fig:mono-lang-treemap} shows the proportions of language families and the largest individual languages in the released data. 

We again emphasize that these web-derived corpora have only undergone essential pre-processing (see above), but no boilerplate removal, fine-grained filtering or extensive cleaning. At the same time, the texts are provided with metadata, which can be employed by end users to conduct their own filtering. 

The datasets come as compressed \texttt{JSONlines} files, where each line is a valid JSON value representing a full document with metadata: %For example:

{\small
\begin{verbatim}
{"id":1, "document_lang":"en",
"scores":["0.76","0.70"],
"langs":["en","en"],
"text":"this is paragraph1\nparagraph2",
"url":"url1", "collection":"collection1"
}
\end{verbatim}
}

The document text is in the `text' field; paragraph-like segments are concatenated using newline separators (here, we have two paragraphs). The `langs' and `scores' fields contain lists with one entry per paragraph. The first corresponds to the paragraph languages identified by FastSpell (both paragraphs are in English here), and the second corresponds to the Monocleaner fluency score of each paragraph (in this case, the first paragraph is slightly closer to `regular' English than the second one). The `url' field provides the original URL from where the document was downloaded, and the `collection' field features the identifier of a specific web crawl where the document was found (for example, `WIDE16'). 
\newpage
\subsubsection{Cleaned Version}
In addition to the de-duplicated version of the monolingual datasets (v1.1), we also published the so called `cleaned' version (v1.2).\footnote{\url{https://hplt-project.org/datasets/v1.2}} In it, we removed full documents which satisfied at least one of the following 5 criteria: 
\begin{enumerate}
    \item URL is in the UT1 blocklist of adult sites.\footnote{\url{https://dsi.ut-capitole.fr/blacklists/}}
    \item less than 5 words per segment (line) on average.
    \item less than 200 characters in the document.
    \item less than 5 segments (lines) in the document.
    \item less than 20\% of the segments in the document share the language identified at document level.
\end{enumerate}

Cleaning further reduced the monolingual dataset size from 11 TB in the de-duplicated version to 8.4 TB. However, we believe the cleaned version is even better suited for training large language models.

% \subsubsection{Crawl Overlap}
% \label{sec:crawl-overlap}
% \textcolor{red}{TODO} Insert statistics on overlaps with CC, to prove we are not simply repeating the existing web corpora.

\begin{figure}
    \centering
    \includegraphics[width=\linewidth]{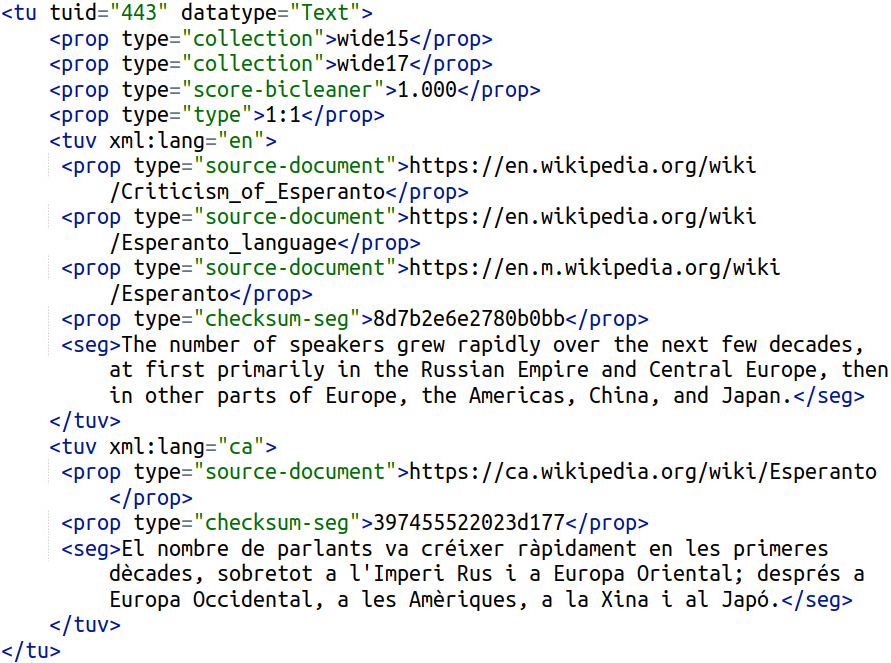}
    \caption{TMX structure of the bilingual datasets.}
    \label{fig:tmux-structure}
\end{figure}

\subsection{Parallel Datasets}
% Note: use zh code for Chinese 1) to keep consistent with monolingual data, 2) other languages didn't specify writing scripts, 3) maybe add zh-Hans in the future?
Our parallel collection is English-centric, with every language paired with English, and includes 18 language pairs with the following languages: Albanian (sq), Arabic (ar), Basque (eu), Bosnian (bs), Catalan (ca), Chinese (zh)\footnote{This release focuses on Chinese written in the traditional script.}, Croatian (hr), Estonian (et), Finnish (fi), Gaelic (ga), Galician (gl), Hindi (hi), Icelandic (is), Macedonian (mk), Maltese (mt), Norwegian Nynorsk (nn), Serbian (sr), and Swahili (sw). We have focused on a wide range of languages in terms of availability and language families.

%: high-resource (ar), mid-resource (ca, fi) and low-resource (ga, is); and in terms of language families: Afro-Asiatic (ar, mt), Bantu (sw), Balto-Slavic (bs, hr, mk, sr), Celtic (ga), Finno-Ugric (fi, et), Indo-Aryan (hi), North Germanic (nn, is), Sino-Tibetan (zh), Romance (ca, gl), and Basque as an isolate.

The data is released in both bitext and TMX format, with the following metadata for each sentence pair: source crawl collection, Bicleaner AI score and, for each segment, the source url(s) and a hash value. An example is depicted in Figure~\ref{fig:tmux-structure}.

\begin{figure*}[t!]
    \centering
    \includegraphics[width=\textwidth]{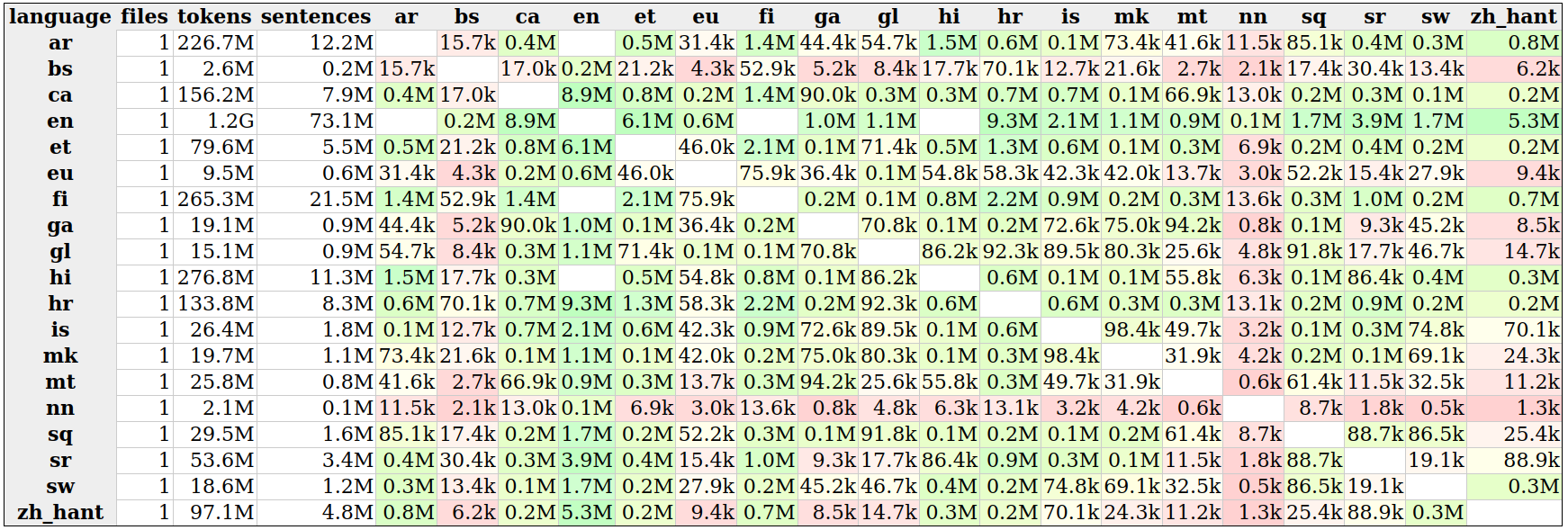}
    \captionsetup{skip=5pt}  % Adjust the skip length as needed
    \caption{Available segments per language pair obtained via pivoting through English taken from the OPUS website. The color scale reflects the size and the counts above the diagonal refer to translation units in TMX files and below refer to aligned bitext segments, which in this case should be identical.}
    \label{fig:pivots}
\end{figure*}

\normalsize
\subsubsection{Statistics}
Statistics for the data release are shown in Table~\ref{tab:extracted_bitexts} to report about parallel segments and  English-side tokens per language pair without filtering (Raw), after processing with Bicleaner AI (Filtered), and after de-duplication (De-duplicated). Raw alignments are also released to enable research on other cleaning methods or quality thresholds.

The parallel corpus contains over 96 million clean and unique sentence pairs and covers over 1.4 billion English tokens. As expected when dealing with low-resource languages, the individual corpus sizes are greatly skewed, with the top five languages accounting for 75\% of the data. The average English sentence length is 14.7 whitespace-separated tokens.

The largest parallel corpora are for Finnish, followed by Arabic and Hindi. While Arabic and Hindi have a large amount of speakers (over 100 million), Finnish is far less represented on the web. The MT model used for Finnish translation is an already existing OPUS-MT model which retrieved a considerably higher number of raw aligned sentences (3.5 billion) compared to other language pairs. For Arabic and Hindi, we experimented with MT systems that were trained on data explicitly avoiding web-crawled content. Whether this approach produces a smaller, but higher quality, set of parallel candidates is still to be investigated.% an unresolved question that needs to be further explored.

Data filtering is an essential step, particularly when handling web-crawled data. We observe a 90\% decrease in size when comparing the raw data with the filtered one. We also apply de-duplication, which further reduces the size by a substantial proportion, it eliminates 90\% of the remaining 10\%. %eliminates a comparable percentage of data.

\vspace{-8pt}

\subsubsection{OPUS Overlaps}
%As demonstrated above in section \ref{sec:crawl-overlap},
Since our parallel data is generated from the same sources as our monolingual release, which contains previously unreleased crawls, we hypothesize they also provide new information. To further investigate this issue, we have computed segment-level overlaps for each language pair with all the existing datasets in OPUS by looking at matching sentence pairs. We release detailed results for this analysis on GitHub.\footnote{\url{https://github.com/Helsinki-NLP/OPUS/tree/hplt2023/corpus/HPLT/v1/overlaps}} On average, only 3.35\% of our data already exists in CCMatrix and 15.72\% in ParaCrawl, two of the most widely used multilingual web-crawled collections.

\begin{comment}
\begin{figure}
    \centering
    \includegraphics[width=\linewidth]{images/overlaps.pdf}
    \caption{Overlaps of our parallel datasets with two other web-crawled corpora, CCMatrix and ParaCrawl. Here we report the summation of all parallel segments only for the language pairs of our datasets.}
    \label{fig:overlaps}
\end{figure}
\end{comment}

\vspace{-8pt}

\subsubsection{MultiHPLT}
Although we plan to specifically target non-English-centric language pairs in the future, for this current release we further leverage resources by pivoting through English, creating a synthetic parallel corpus that includes all possible language pair combinations, multiHPLT. We report the statistics of those additional 
% the artificial 
resources in Figure~\ref{fig:pivots}.
It covers 171 language pairs and over 157 million parallel sentences.

\section{Conclusions and Future Work}
\label{sec:conclusions}

In this paper, we have introduced the HPLT language resources, a new collection of monolingual and bilingual corpora leveraging data from web crawls and released under the permissive \texttt{CC0} license. We have focused on curating data for medium- to low-resource languages with the hope that we will encourage the development of
technology in these languages. One of our contributions is the extraction of massive multilingual text resources from Internet Archive crawls, which we have shown to offer data not found in other web-based corpora. Our data releases are, to our best knowledge, the largest \textit{fully accessible} multilingual text corpora ever released.%\footnote{\url{https://hplt-project.org/datasets/}}
% to contain more raw text inside each url, since it has deeper crawls.

While the resources presented in this paper mark a significant milestone, there are several avenues for future research. We plan to expand the language coverage and include non--English-centric language pairs; also to enrich the datasets with further metadata.
%Regarding data quality, the ultimate measure is human evaluation, and in future releases, we will seek to include human-contributed quality scores at the segment level. 
Additionally, we also work to improve our tooling and, correspondingly, corpus quality. This includes deploying better language identification, %in \texttt{warc2text}
tackling boilerplate identification and exploring the feasibility and benefit of performing bitexting across shards. We also seek to make better use of our HPC resources through additional automation of our data pipeline, and stability improvements for our AMD ports of MarianNMT and  Megatron-DeepSpeed. In future releases of the project, we will include LLMs and MT models across our supported language set, as well as the training pipelines used to create them. 

Finally, our main goal is to contribute to the NLP research field by providing massive high-quality datasets, therefore we take this opportunity to call for action and ask the community to contribute both raw data sources and processed corpora so that we can include them to our collection.

\section{Environmental Considerations} 
Developing large-scale datasets for language modeling is an expensive task that has an environmental impact. Releasing the datasets publicly in open-source repositories directly reduces this impact, as they can be reused instead of creating from scratch. However, we believe it's important to keep track of and share how much carbon is produced when building these large datasets. Next we report our estimates in hours:
\begin{itemize}
    \item Data download:  62K CPU
    \item Data processing: 72K CPU
    \item Monotexting: 800K CPU
    \item Bitexting: 2,2M CPU and 53K GPU
\end{itemize}
The total amount of hours spent would be roughly 5M CPU hours and 50K GPU hours. Note that the most expensive task is generating the parallel corpora since it involves translating all documents into English. Note also that the LUMI supercomputer uses renewable, carbon-neutral energy.\footnote{\url{https://lumi-supercomputer.eu/sustainable-future/}}

\section{Limitations}

In this paper, we focus on the description of the construction of the first release of the HPLT language resources. We are aware that we do not provide a qualitative analysis of the datasets, and we do not train models to validate the quality of the data. While we plan to do this, these experiments are complex and expensive and a comprehensive evaluation falls out of the scope of the paper, with its main goal being to present and describe the datasets.

\section{Acknowledgements}
This project has received funding from the European Union’s Horizon Europe research and innovation programme under Grant agreement No 101070350 and from UK Research and Innovation (UKRI) under the UK government’s Horizon Europe funding guarantee [grant number 10052546]. The contents of this publication are the sole responsibility of its authors and do not necessarily reflect the opinion of the European Union.
The authors wish to thank CSC - IT Center for Science, Finland for computational resources and support.

\section{Bibliographical References}\label{sec:reference}

\bibliographystyle{lrec-coling2024-natbib}
\bibliography{lrec-coling2024-example}

\section{Language Resource References}
\label{lr:ref}
\bibliographystylelanguageresource{lrec-coling2024-natbib}
\bibliographylanguageresource{languageresource}

\section*{\textit{Appendix A. Source Crawl Distribution per Language }}
\label{sec:appendixa}

\begin{figure}[ht!]
\centering
\includegraphics[width=\linewidth]{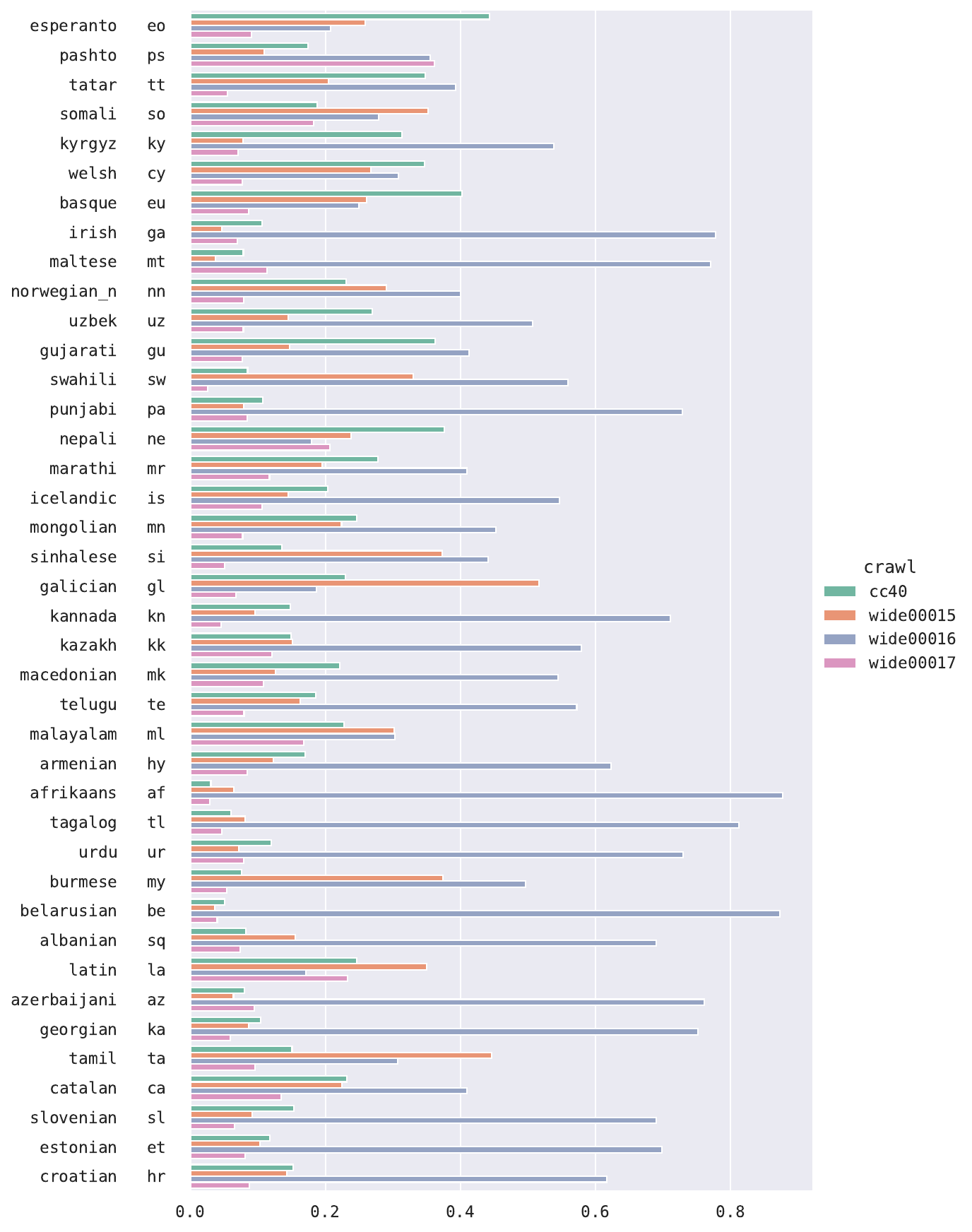}
\caption{Proportions of text volume in bytes coming from each crawl, part 1.}
\label{fig:text_bytes_props1}
\end{figure}

\begin{figure}[ht!]
\centering
\includegraphics[width=\linewidth]{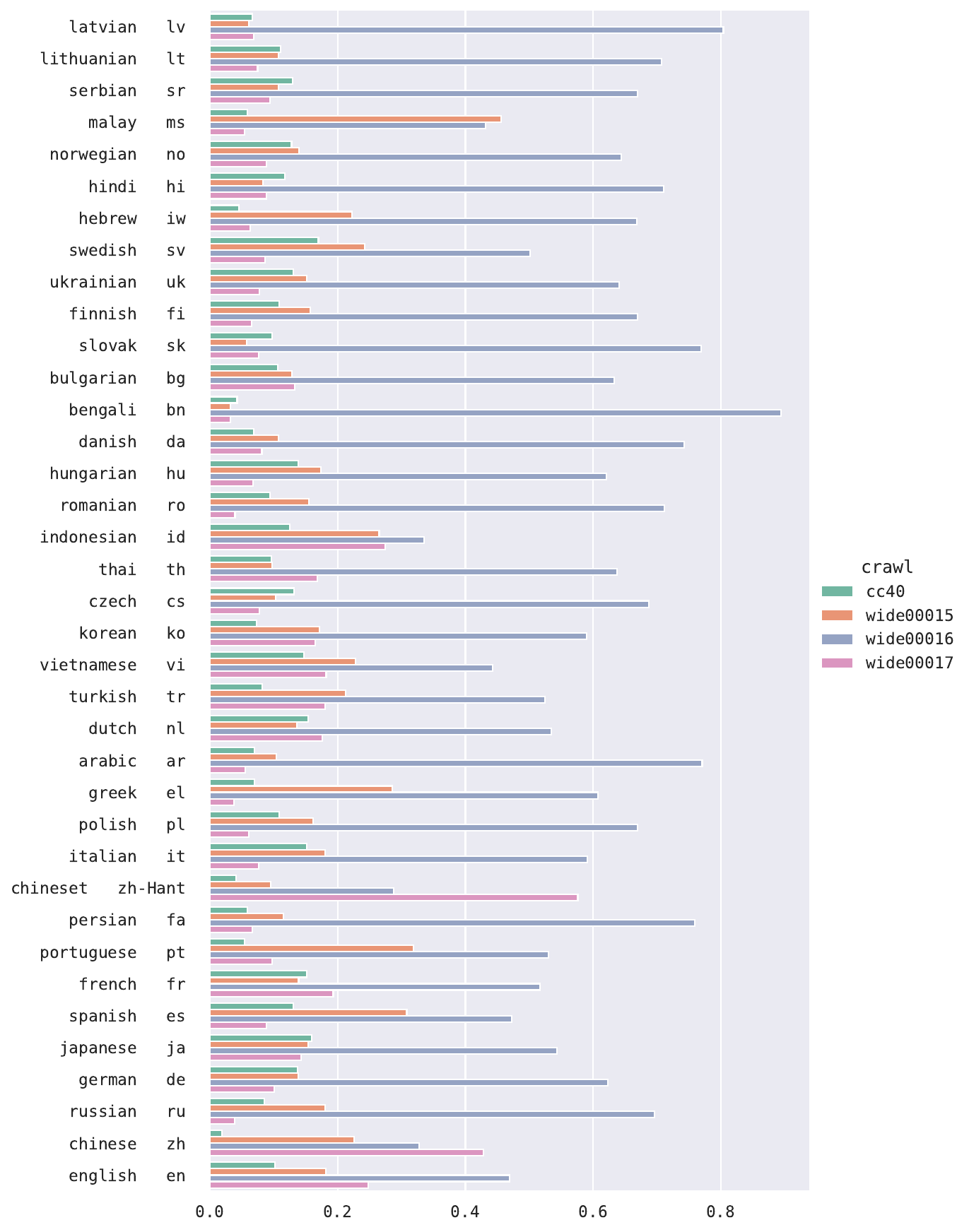}
\caption{Proportions of text volume in bytes coming from each crawl, part 2.}
\label{fig:text_bytes_props2}
\end{figure}

\section*{\textit{Appendix B. Per-Language Monolingual Statistics}}
\label{sec:language_stats}

Table~\ref{tab:deduped_release_perlang_stats} shows the total sizes of texts in each language for the deduplicated publicly released data. The number of segments (lines), words and bytes are as reported by the Unix \texttt{wc(1)} tool, see its documentation for definitions of a line and a word. The volume of texts significantly ranges from 1.0 GB for text classified by CLD2 as Esperanto to 20.3 TB for English, while the number of documents has the minimum of 143K for Pashto and the maximum of 1.8B for English. We foresee a high percentage of documents mis-classified  by CLD2 due to the huge amount of noisy data that it receives at this stage.

\begin{scriptsize}
\clearpage
\onecolumn
\begin{longtable}{ll|rrrrr}
\toprule
\textbf{Language} & \textbf{Code} & \textbf{\# Segments} & \textbf{\# Words} & \textbf{Characters} & \textbf{\# Bytes} & \textbf{\# Documents} \\
\midrule

Esperanto           & eo            & 2.29e+07                & 1.47e+08               & 9.64e+08             & 9.91e+08             & 1.77e+05      \\
Pashto              & ps            & 2.65e+07                & 1.68e+08               & 8.22e+08             & 1.32e+09             & 1.43e+05      \\
Tatar               & tt            & 2.64e+07                & 1.34e+08               & 9.65e+08             & 1.63e+09             & 1.72e+05      \\
Welsh               & cy            & 4.76e+07                & 2.44e+08               & 1.63e+09             & 1.64e+09             & 2.85e+05      \\
Kyrgyz              & ky            & 2.83e+07                & 1.46e+08               & 1.12e+09             & 1.96e+09             & 1.88e+05      \\
Somali              & so            & 4.41e+07                & 3.00e+08               & 1.99e+09             & 2.01e+09             & 3.75e+05      \\
Irish               & ga            & 1.24e+08                & 5.20e+08               & 3.38e+09             & 3.61e+09             & 9.32e+05      \\
Norwegian           & nn            & 1.41e+08                & 6.40e+08               & 4.36e+09             & 4.46e+09             & 7.53e+05      \\
Basque              & eu            & 1.46e+08                & 7.17e+08               & 5.23e+09             & 5.30e+09             & 1.01e+06      \\
Swahili             & sw            & 1.55e+08                & 9.11e+08               & 5.88e+09             & 5.94e+09             & 9.84e+05      \\
Maltese             & mt            & 1.69e+08                & 8.19e+08               & 5.69e+09             & 6.01e+09             & 4.84e+05      \\
Gujarati            & gu            & 8.39e+07                & 4.73e+08               & 2.95e+09             & 6.10e+09             & 4.55e+05      \\
Uzbek               & uz            & 9.27e+07                & 5.56e+08               & 4.38e+09             & 6.12e+09             & 6.33e+05      \\
Punjabi             & pa            & 1.23e+08                & 5.33e+08               & 3.06e+09             & 6.41e+09             & 8.88e+05      \\
Galician            & gl            & 2.22e+08                & 1.29e+09               & 8.34e+09             & 8.56e+09             & 1.79e+06      \\
Kannada             & kn            & 1.67e+08                & 5.78e+08               & 4.07e+09             & 8.71e+09             & 5.58e+05      \\
Icelandic           & is            & 3.23e+08                & 1.34e+09               & 9.16e+09             & 9.99e+09             & 1.44e+06      \\
Tagalog             & tl            & 2.40e+08                & 1.63e+09               & 1.15e+10             & 1.16e+10             & 1.20e+06      \\
Sinhalese           & si            & 1.28e+08                & 9.18e+08               & 5.82e+09             & 1.19e+10             & 5.64e+05      \\
Macedonian          & mk            & 2.58e+08                & 1.11e+09               & 7.31e+09             & 1.20e+10             & 1.25e+06      \\
Mongolian           & mn            & 1.86e+08                & 1.09e+09               & 7.55e+09             & 1.27e+10             & 1.06e+06      \\
Marathi             & mr            & 1.58e+08                & 9.12e+08               & 6.06e+09             & 1.31e+10             & 8.57e+05      \\
Afrikaans           & af            & 4.48e+08                & 1.87e+09               & 1.33e+10             & 1.35e+10             & 1.37e+06      \\
Kazakh              & kk            & 2.31e+08                & 1.01e+09               & 7.77e+09             & 1.37e+10             & 1.43e+06      \\
Armenian            & hy            & 3.55e+08                & 1.31e+09               & 9.47e+09             & 1.54e+10             & 1.36e+06      \\
Nepali              & ne            & 1.65e+08                & 1.06e+09               & 6.86e+09             & 1.56e+10             & 1.36e+06      \\
Telugu              & te            & 2.28e+08                & 1.06e+09               & 7.56e+09             & 1.59e+10             & 1.61e+06      \\
Urdu                & ur            & 2.68e+08                & 2.02e+09               & 1.11e+10             & 1.61e+10             & 2.23e+06      \\
Belarusian          & be            & 2.92e+08                & 1.39e+09               & 1.14e+10             & 1.93e+10             & 1.26e+06      \\
Malayalam           & ml            & 2.11e+08                & 1.05e+09               & 8.75e+09             & 1.94e+10             & 1.13e+06      \\
Burmese             & my            & 3.00e+08                & 1.25e+09               & 9.81e+09             & 1.97e+10             & 8.26e+05      \\
Latin               & la            & 5.76e+08                & 3.32e+09               & 2.40e+10             & 2.42e+10             & 4.81e+06      \\
Georgian            & ka            & 5.09e+08                & 1.68e+09               & 1.22e+10             & 2.51e+10             & 1.67e+06      \\
Azerbaijani         & az            & 7.62e+08                & 2.94e+09               & 2.22e+10             & 2.52e+10             & 3.00e+06      \\
Albanian            & sq            & 7.37e+08                & 3.78e+09               & 2.53e+10             & 2.65e+10             & 3.22e+06      \\
Latvian             & lv            & 1.48e+09                & 5.39e+09               & 3.98e+10             & 4.23e+10             & 5.12e+06      \\
Estonian            & et            & 1.59e+09                & 5.85e+09               & 4.48e+10             & 4.60e+10             & 5.84e+06      \\
Slovenian           & sl            & 1.58e+09                & 7.04e+09               & 4.79e+10             & 4.90e+10             & 5.82e+06      \\
Catalan             & ca            & 1.16e+09                & 7.88e+09               & 4.94e+10             & 5.10e+10             & 7.79e+06      \\
Lithuanian          & lt            & 1.71e+09                & 7.78e+09               & 5.40e+10             & 5.67e+10             & 7.40e+06      \\
Tamil               & ta            & 6.31e+08                & 3.87e+09               & 2.95e+10             & 6.58e+10             & 2.47e+06      \\
Norwegian Bokmål    & nb            & 3.19e+09                & 1.39e+10               & 9.21e+10             & 9.41e+10             & 1.46e+07      \\
Slovak              & sk            & 3.89e+09                & 1.39e+10               & 9.50e+10             & 1.02e+11             & 1.40e+07      \\
Malay               & ms            & 3.25e+09                & 1.65e+10               & 1.00e+11             & 1.02e+11             & 8.36e+06      \\
Bengali             & bn            & 2.43e+09                & 8.23e+09               & 5.51e+10             & 1.29e+11             & 5.97e+06      \\
Finnish             & fi            & 4.76e+09                & 1.69e+10               & 1.38e+11             & 1.42e+11             & 1.95e+07      \\
Serbo-Croatian      & hbs           & 4.77e+09                & 1.94e+10               & 1.32e+11             & 1.45e+11             & 1.78e+07      \\
Hebrew              & he            & 3.73e+09                & 1.55e+10               & 9.27e+10             & 1.52e+11             & 1.12e+07      \\
Danish              & da            & 4.58e+09                & 2.21e+10               & 1.53e+11             & 1.56e+11             & 2.36e+07      \\
Hindi               & hi            & 2.77e+09                & 1.37e+10               & 8.12e+10             & 1.62e+11             & 1.14e+07      \\
Bulgarian           & bg            & 3.51e+09                & 1.55e+10               & 1.04e+11             & 1.73e+11             & 1.33e+07      \\
Swedish             & sv            & 6.56e+09                & 2.83e+10               & 1.88e+11             & 1.95e+11             & 3.00e+07      \\
Hungarian           & hu            & 6.89e+09                & 2.65e+10               & 1.99e+11             & 2.17e+11             & 2.85e+07      \\
Ukrainian           & uk            & 3.18e+09                & 1.82e+10               & 1.34e+11             & 2.31e+11             & 1.79e+07      \\
Romanian            & ro            & 7.07e+09                & 3.28e+10               & 2.41e+11             & 2.47e+11             & 2.49e+07      \\
Korean              & ko            & 9.55e+09                & 2.90e+10               & 1.49e+11             & 2.80e+11             & 4.45e+07      \\
Czech               & cs            & 9.88e+09                & 4.10e+10               & 2.62e+11             & 2.87e+11             & 3.86e+07      \\
Vietnamese          & vi            & 9.49e+09                & 6.50e+10               & 3.17e+11             & 3.92e+11             & 4.01e+07      \\
Thai                & th            & 8.43e+09                & 2.20e+10               & 1.93e+11             & 4.05e+11             & 2.95e+07      \\
Indonesian          & id            & 9.66e+09                & 6.92e+10               & 4.81e+11             & 4.84e+11             & 4.58e+07      \\
Turkish             & tr            & 1.03e+10                & 6.49e+10               & 4.55e+11             & 4.93e+11             & 5.94e+07      \\
Dutch               & nl            & 1.65e+10                & 7.18e+10               & 5.18e+11             & 5.23e+11             & 6.66e+07      \\
Arabic              & ar            & 9.20e+09                & 5.15e+10               & 3.23e+11             & 5.27e+11             & 4.66e+07      \\
Greek               & el            & 1.13e+10                & 5.22e+10               & 3.40e+11             & 5.47e+11             & 3.06e+07      \\
Polish              & pl            & 1.93e+10                & 8.54e+10               & 5.95e+11             & 6.17e+11             & 8.29e+07      \\
Persian             & fa            & 1.34e+10                & 7.04e+10               & 3.84e+11             & 6.45e+11             & 4.23e+07      \\
Italian             & it            & 2.27e+10                & 1.14e+11               & 7.68e+11             & 7.77e+11             & 9.65e+07      \\
Portuguese          & pt            & 2.74e+10                & 1.32e+11               & 8.27e+11             & 8.53e+11             & 1.04e+08      \\
French              & fr            & 4.36e+10                & 2.14e+11               & 1.37e+12             & 1.41e+12             & 1.76e+08      \\
German              & de            & 4.35e+10                & 1.93e+11               & 1.43e+12             & 1.46e+12             & 2.26e+08      \\
Spanish             & es            & 4.45e+10                & 2.50e+11               & 1.56e+12             & 1.60e+12             & 2.01e+08      \\
Japanese            & ja            & 5.14e+10                & 9.14e+10               & 8.92e+11             & 1.98e+12             & 2.19e+08      \\
Russian             & ru            & 1.14e+11                & 4.93e+11               & 3.65e+12             & 6.02e+12             & 3.97e+08      \\
Chinese             & zh            & 1.73e+11                & 3.35e+11               & 3.35e+12             & 7.51e+12             & 1.20e+09      \\
English             & en            & 3.87e+11                & 2.86e+12               & 2.03e+13             & 2.03e+13             & 1.78e+09      \\
\midrule
Total        &       & 1.11e+12      & 5.64e+12                & 4.05e+13               & 5.01e+13             & 5.25e+09                         \\
\bottomrule
\caption{Languages in the deduplicated public data release: the number of segments (new line symbols), words (as defined by \texttt{wc(1)}), characters, bytes and documents. Ordered by size in bytes.}
\label{tab:deduped_release_perlang_stats}
\end{longtable}
\end{scriptsize}

\end{document}